\documentclass[Afour,times]{sagej}
\pdfoutput=1
\usepackage{amsmath}
\usepackage{mathtools}
\usepackage{amssymb}
\usepackage{gensymb}
\usepackage[sort&compress,numbers]{natbib}

\usepackage[toc,page]{appendix}
\usepackage{lscape} 
\usepackage{caption}
\usepackage{subcaption}
\usepackage{booktabs}
\usepackage{parskip}
\usepackage{fdsymbol}

\usepackage{changepage}

\usepackage{scalerel}

\usepackage{setspace}
\usepackage[boxruled,linesnumbered]{algorithm2e}
\usepackage{etoolbox}
\usepackage{multicol}
\usepackage{sectsty}
\usepackage{blindtext}
\usepackage{graphicx}

\usepackage{tabularx}   
\usepackage{blindtext}
\usepackage{enumitem}
\usepackage{url}
\usepackage{lineno}

\begin{document}

\runninghead{Poole \emph{et al.}}
\title{On statistic alignment for domain adaptation in structural health monitoring}
\author{Jack Poole\affilnum{1},
 Paul Gardner\affilnum{1},
Nikolaos Dervilis\affilnum{1}, Lawrence Bull\affilnum{2}, Keith Worden \affilnum{1}}
\affiliation{\affilnum{1}Dynamics Research Group, Department of Mechanical Engineering, University of Sheffield, Mappin Street, Sheffield, S1 3JD\\
\affilnum{2}The Alan Turing Institute, British Library, London, NW1 2DB}
\email{jpoole4@sheffield.ac.uk}

\begin{abstract}
The practical application of structural health monitoring (SHM)  is often limited by the availability of labelled data. Transfer learning - specifically in the form of domain adaptation (DA) -- gives rise to the possibility of leveraging information from a \emph{population} of physical or numerical structures, by inferring a mapping that aligns the feature spaces. Typical DA methods rely on nonparametric distance metrics, which require sufficient data to perform density estimation. In addition, these methods can be prone to performance degradation under class imbalance. To address these issues, statistic alignment (SA) is discussed, with a demonstration of how these methods can be made robust to class imbalance, including a special case of class imbalance called a partial DA scenario. SA is demonstrated to facilitate damage localisation with no target labels in a numerical case study, outperforming other state-of-the-art DA methods. It is then shown to be capable of aligning the feature spaces of a real heterogeneous population, the Z24 and KW51 bridges, with only 220 samples used from the KW51 bridge. Finally, in scenarios where more complex mappings are required for knowledge transfer, SA is shown to be a vital pre-processing tool, increasing the performance of established DA methods.
\end{abstract}


\maketitle
\section{Introduction}

The data-driven approach to structural health monitoring (SHM) is often limited by availability of labelled data \cite{FarrarC.R.CharlesR.2013Shm:}. Unsupervised machine learning techniques have been used to detect the presence of damage by utilising outlier analysis  \cite{Dervilis2015}. However, SHM-frameworks that allow for data to be categorised as a specific health-state largely rely on supervised statistical models, preventing their application in practical scenarios. This issue motivates the leveraging of information from a population of structures, giving rise to a new discipline -- \emph{population--based SHM} (PBSHM) \cite{Bull2021,Gosliga2021,Gardner2021}. By considering labelled information across a population of structures, it is more likely that the shared label information would be adequate to facilitate more reliable and detailed diagnostics. 

A critical challenge in PBSHM is that data acquired from different structures will differ in their underlying distributions, invalidating the assumption that the training and testing data were drawn from the same distribution -- an assumption that conventional machine-learning models typically make \cite{bishop}. Variation in the data distributions will occur for a number of reasons; in \emph{homogeneous} populations (i.e. nominally identical structures), manufacturing variations and operating conditions will lead to differences. Populations of structures that cannot be considered nominally identical -- \emph{heterogeneous} populations \cite{Gosliga2021} -- may also differ significantly because of structural differences. One approach is to utilise information from across a population is to infer a mapping that aligns the feature spaces.

\begin{figure}[t!]
     \centering
     \includegraphics[width=0.4\textwidth]{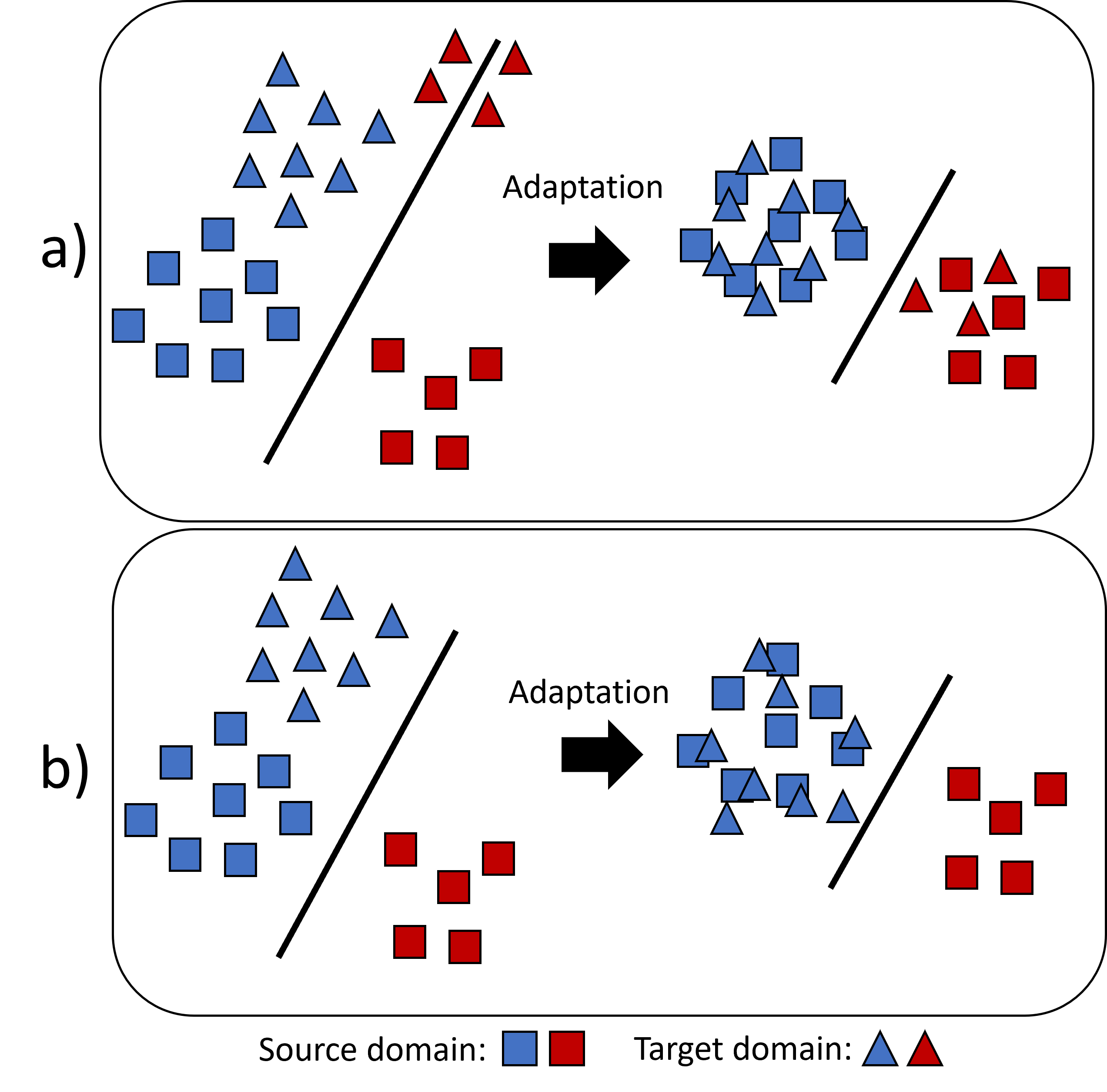}
     \caption{Domain adaptation scenarios showing standard domain adaptation in a and partial domain adaptation in b.}
     \label{fig: cartoon}
\end{figure}

 \emph{Domain adaptation} (DA) -- a sub-field of \emph{transfer learning} (TL) -- aims to achieve this goal by learning a mapping that aligns the distributions of the feature spaces\cite{Zhuang2021}, illustrated in Figure \ref{fig: cartoon} . As such, a statistical model that will generalise well to a target structure (target domain), can be learnt by using data from a source structure (source domain). Typically, these technologies are based on distribution distance metrics, which learn a mapping via nonlinear kernels \cite{JialinPan2011,Long2013,Long2014,Wang2017} or deep neural networks \cite{Long2015,Ho2002,Yeh2016,tzeng2017adversarial,Long2017,Long2018, Zhao2018}. Other approaches align subspaces, assuming that they lie on a common manifold \cite{Gong2012}. 

These technologies have the potential to advance SHM and facilitate more informative diagnostics. Chakraborty \emph{et al.} used transfer learning to address the issue of sensor coverage in a fatigue damage case \cite{chakraborty2011structural}. This application differs from the current paper, which aims to transfer information between structures. A large portion of the literature focusses on deep neural network (DNN) based DA, notably Xu \emph{et al.} use a physics-informed deep DA approach to align the frequency response of multiple structures to perform damage detection and quantification \cite{Xu2020}.  There have also been a number of attempts to apply deep-DA to perform fault diagnosis in machines under changing loading conditions and rotation speeds \cite{wang2020triplet,Li2019,Michau2019,Li2020, jiao2020residual, ding2021remaining}. These works focus on end-to-end DNNs (often convolutional neural networks), with high-dimensional frequency responses used as features; thus, because of the curse of dimensionality \cite{bishop} and the large number of parameters in these networks, large quantities of data are required to train these methods such that they generalise well. Conversely, this paper aims to develop an approach to DA that can be applied with limited data, as engineering datasets, particularly SHM datasets, are often sparse and data are expensive to obtain.  Another approach that has been investigated focuses on fine-tuning DNNs for image-based crack detection \cite{Gao2018, dorafshan2018comparison}. This approach differs from unsupervised DA, as it does not align the domains and  requires target labels.  

DA has also been demonstrated in a PBSHM setting, Gardner \emph{et al.} have shown that DA can be used to transfer localisation labels between numerical and experimental structures \cite{Gardner2020}, two heterogeneous aircraft wings \cite{gardner2022population}, and between pre- and post-repair states in aircraft wings \cite{Gardner2020b}. In Bull \emph{et al.} a population of six experimental tailplanes was used to demonstrate transferring a damage detector \cite{bull2021transfer}. These works have largely focused on methods that rely on kernel-based nonparametric density estimation. Nonparametric density estimation is known to require large quantities of data to provide accurate estimates \cite{Song2013}, which may limit these DA approaches in practical SHM scenarios. 

In addition, the existing approaches do not address issues relating to class imbalance and partial domain adaptation. Class imbalance, where the data quantity for each class is inconsistent, is inevitable in SHM datasets because of the rarity of damage and spurious environmental conditions \cite{mao2021toward}. A specific class imbalance scenario in TL called \emph{partial DA} \cite{Cao2018a}, refers to the scenario where the target domain contains data pertaining to fewer classes than the source, demonstrated in Figure \ref{fig: cartoon}. It is crucial that DA methods that are robust to these issues are developed since previously used methods are prone to performance degradation in these scenarios \cite{Wang2017a,Cao2018a}. For fault detection in machines, instance-weighting has been used to achieve impressive classification improvement for partial DA \cite{jiao2019classifier,deng2021double}. However, these approaches also rely on DNNs and to the best of the authors knowledge partial DA remains uninvestigated in the context of SHM.

The main idea of this paper is that \emph{statistic alignment} (SA) \cite{Zhang2017}, a branch of DA that directly aligns the lower-order statistics, can be used as a low-risk form of DA in scenarios with limited data availability and engineering insight can be used to adapt SA to address the issues with class imbalance and partial DA. The main contributions of this paper are as follows:
\begin{enumerate}
    \item SA is shown to perform effective knowledge transfer where standard SHM, and the major branches of DA methods utilised in prior work (manifold, kernel and DNNs) are ineffective when standard practices for standardisation are followed. 
    \item An approach to SA utilising only normal-condition data that is robust to class imbalance and partial DA scenarios is proposed, along with two methods: normal-condition alignment (NCA) and normal correlation alignment (NCORAL). These methods are shown capable of adapting both numerical (shear structures) and real heterogeneous populations (the Z24 and KW51 bridges), demonstrating that this approach could facilitate the use of supervised or semi-supervised machine learning methods before any damage has occurred in the target structure.
    \item The limitations of SA are discussed; a case study demonstrates that, when the higher-order statistics differ significantly, further DA can be beneficial. It is found that in this scenario, each of the major branches of DA fail to adapt the domains following standard practice for pre-processing, but they succeed following SA, suggesting that SA is an essential pre-processing step for the use of these methods.  

\end{enumerate}

\begin{figure*}[h!]
     \centering
     \includegraphics[width=0.8\textwidth]{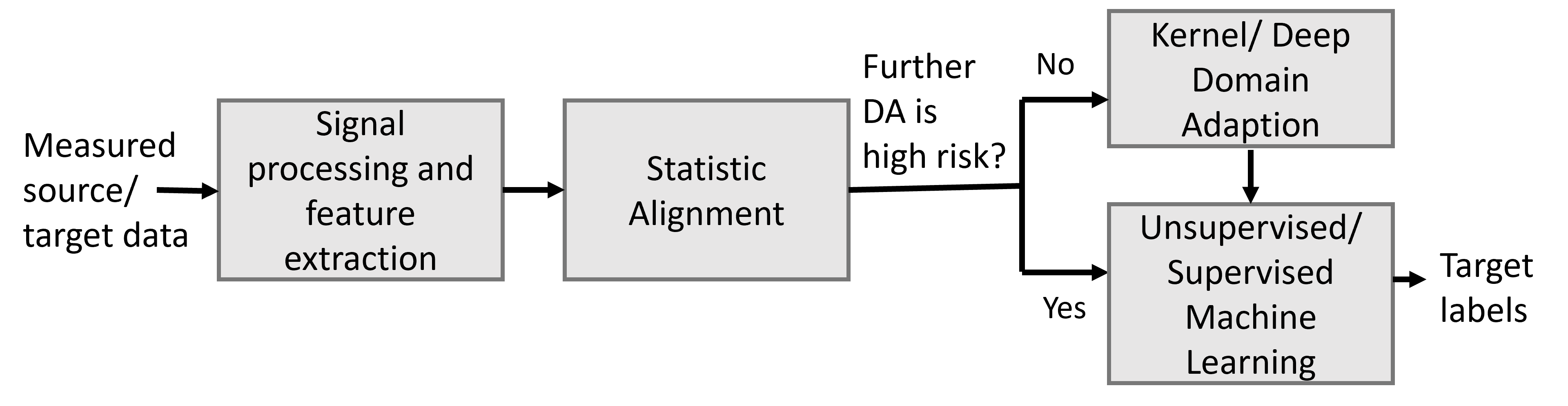}
     \caption{The proposed flow for domain adaptation in PBSHM. Note: some DA methods may incorporate domain adaptation and a predictive function (i.e. a classifier) in a single model.}
     \label{fig: flow}
\end{figure*}

The proposed approach to DA in PBSHM is given in Figure \ref{fig: flow}. The results in this paper suggest that SA should be used as initial adaptation, after which a statistical model can be learnt directly on the source domain, or further DA could improve adaption prior learning a statistical model. When applying DA, there is a risk that performance is worse than using the target alone \cite{Zhang2020}; thus, this paper discusses why SA is a low-risk form of DA in low data scenarios and when further DA may be appropriate.

The outline of this paper is as follows. In section 2, the necessary  background is given on DA and SA. The disadvantages of conventional DA are discussed in regard to data availability, class imbalance and partial DA, and the proposed approach that aligns the domains using only normal-condition data is introduced. Section 3 demonstrates that SA can transfer label information between numerical structures in both conventional DA and partial DA, where  previously investigated DA methods fail, and it is shown that the proposed methods are particularly robust for partial DA. In Section 4, the application of NCORAL is discussed for a real population bridges, the Z24 \cite{maeck2003description} and KW51 bridges \cite{maes2021monitoring}. This population presents two partial DA scenarios, which include three domains pertaining to the two heterogeneous bridges, as well as, a pre- and post-repair state in the KW51 bridge dataset. Section 5 discusses the limitations of only aligning the lower-order statistics and introduces SA as a pre-processing step for other prominent DA algorithms, with another numerical case study suggesting that SA may be an essential pre-processing step for the application of many DA algorithms. Finally, Section 6 presents a discussion on SA and discusses future work.  A GitHub repository accompanies this paper: https://github.com/Jack-Poole/Statistic-Alignment.git.

\section{Domain Adaptation}

Before defining DA, it is important to introduce two key objects in transfer learning, a domain and task \cite{Zhuang2021}:
\begin{itemize}
    \item A \emph{domain} $\mathcal{D}=\{\mathcal{X}, P(X)\}$, defined by of a feature space $\mathcal{X}$ and a marginal probability distribution $P(X)$.
    \item A \emph{task} for a given domain is defined by $\mathcal{T}=\{\mathcal{Y},f(\cdot)\}$, where $\mathcal{Y}$ is the label space and $f(\cdot)$ is a predictive function learnt from a finite sample  $\{\mathbf{x}_{i},y_{i}\}_{i=1}^{n}$, where $x_i \in \mathcal{X}$ and $y_i \in \mathcal{Y}$.
\end{itemize}

In \emph{unsupervised domain adaptation}, a source domain $\mathcal{D}_s=\{\mathbf{x}_{s,i},y_{s,i}\}_{i=1}^{n_s}$, with $n_s$ source instances $\mathbf{x}_{s,i}$, each with labels $y_{s,i}$ and a target domain $\mathcal{D}_t=\{\mathbf{x}_{t,j}\}_{j=1}^{n_t}$ with $n_t$ unlabelled target instances, $\mathbf{x}_{t,j}$, are used to learn a classifier that generalises to the target domain. It is assumed that there are differences in the marginal distributions $P(X_s) \neq P(X_t)$, and/or the conditional distributions $P(\mathbf{y}_s|X_s) \neq P(\mathbf{y}_t|X_t)$. Thus, DA aims to find a mapping that aligns the data distributions. The focus of this paper is homogeneous DA, where the feature spaces are assumed the same $\mathcal{X}_s=\mathcal{X}_t$. 

This paper is also concerned with \emph{partial DA} problems, where the available target data pertains to fewer classes than the source, i.e. the target label space is a subset of the source $\mathcal{Y}_t \subset \mathcal{Y}_s$ \cite{Zhuang2021}. For example, consider a discrete damage localisation problem with five locations, the source may have data for all five locations, whereas the target could only have data pertaining to one location, where the aim of using partial DA would be to use the data in the source domain to learn what location the damage in the target relates to. 

A major consideration when attempting unsupervised DA is that the lack of labels in the target domain makes validation challenging. Reducing the risk of \emph{negative transfer} \cite{Zhang2020}, where DA causes performance degradation, is an ongoing focus of TL research, and is of particular importance in SHM where safety critical and high cost assets are at risk.
\vspace{-0.3cm}

\subsection{Statistic Alignment} 

Domain adaptation methods typically attempt to completely harmonise the data distributions, often utilising nonparametric distance metrics to find a nonlinear mapping  \cite{Zhuang2021}. Generally, these metrics rely on the data in both domains being sufficient to perform accurate density estimation. However, in SHM problems, data from a given structure are likely to be limited, particularly for more informative data, such as damage-states or disparate environmental conditions. A branch of DA, \emph{statistic alignment}, provides an alternative solution to align the datasets; and it is defined as:

\textbf{Definition 1}: \emph{Statistic alignment} (SA) methods for DA directly align the lower-order statistics of the domains via affine transformations and scaling.

These approaches typically focus on matching the first- and second-order statistics, and are most appropriate when:
\begin{itemize}
    \item Data can be assumed Gaussian, which is a common assumption in SHM for the linear response of a structure \cite{FarrarC.R.CharlesR.2013Shm:}. 
    \item The higher-order statistics are already similar.
    \item Data are limited as robustly estimating the lower-order statistics requires significantly less data than nonparametric density estimation.
\end{itemize}

In addition to being less data intensive, SA could also facilitate better visualisation of the domain data. Many prominent DA methods \cite{Zhuang2021,Weiss2016} project the data into a latent space via a nonlinear mapping. In comparison, SA maintains the structure of the original feature space, as it is limited to affine transformations; this can be useful for physically-interpretable features, which are common in SHM, because features often correspond to some physical process; for example, an increase in natural frequency can be interpreted as a stiffening effect. 

One of the most popular SA methods is \emph{correlation alignment} (CORAL) \cite{Sun}, which aligns the source correlation with the target. The multiple outlook mapping algorithm (MOMAP) is a similar approach that aligns the principal components \cite{Harel2011}. He \emph{et al.} proposed Euclidean alignment (EA), which aligns the means of the covariance matrices for 2D electroencephalogram (EEG) features, and demonstrated that SA methods can be generalised to cases with multiple sources \cite{He2020}.

A related branch of DA concerns batch normalisation approaches \cite{Li2018,Chang2019,Seo2020}, which align the means and standard deviations of the \emph{activations} in deep neural networks. These methods differ to SA, as the statistics of the original data are not directly aligned and the objective of the network (classification etc.) will influence the feature representation. 

\subsubsection{Standardisation as Statistic Alignment}

Standardisation, a form of normalisation, is commonly used in conventional
machine learning to give each feature equal treatment \cite{bishop}. In DA, it also has the potential to align the means $\mu$ and standard deviations $\sigma$ of the marginal distributions $P(X_s)$ and $P(X_t)$ in an unsupervised manner, with the following transformations,

\begin{equation}
\begin{aligned}
     \mathbf{z}_s^{(j)}=\frac{\mathbf{x}_s^{(j)}-\boldsymbol{\mu}_{s} }{\boldsymbol{\sigma}_{s}}
\end{aligned}\label{eq: source stand}
\end{equation}

\begin{equation}
     \mathbf{z}_t^{(j)}=\frac{\mathbf{x}_t^{(j)}-\boldsymbol{\mu}_{t} }{\boldsymbol{\sigma}_{t}}
\label{eq: target stand}\end{equation}

\noindent where $\boldsymbol{\mu}_{s}$ and $\boldsymbol{\mu}_{t}$ are the means of the source and target; $\boldsymbol{\sigma}_{s}$ and $\boldsymbol{\sigma}_{t}$ are the respective standard deviations. In this paper, this form of standardisation will be referred to as \emph{A-standardisation}. On the other hand, standard practices for machine learning would suggest that the statistics applied to all data should be the same; for example, if some data from the source and target domain are considered training data the statistics would be calculated from $X= X_s\cup  X_t$, which may remove the effect of measurement scale without changing relative mean distance or scale between the domains. This method will be called \emph{N-standardisation} in this paper.

Previous work has studied various approaches to normalisation to improve the performance of a popular DA algorithm, transfer component analysis (TCA) \cite{Nam2013}, showing standardisation can lead to better adaptation. However, it suggests a range of approaches, including methods that would not align the statistics (such as N-standardisation), and does not consider using standardisation alone to perform adaptation. The current paper first considers standardisation as a form of domain adaptation itself, before considering it as a pre-processing method.

\subsubsection{Correlation Alignment}

A-standardisation aligns the domains, ignoring the correlation between features. Correlation Alignment (CORAL) \cite{Sun}, extends this method to also align the covaraince. This is achieved by transforming the source domain via a linear transformation matrix $A$, such that,
\begin{equation}
    \underset{A}{min}||\hat{C}_s-C_t||_F^2=\underset{A}{min}||A^TC_sA-C_t||_F^2
\end{equation}
\noindent where $\hat{C}_s$ is the covariance of the transformed source, $C_s$ and $C_t$ denote the covariance matrices of the source and target respectively, and $||\cdot||_F$ is the Frobenius norm. In many SHM cases, CORAL may improve generalisation of the source classifier, but estimating covariance suffers from the curse of dimensionality. If the number of observations in $X_s$ or $X_t$ are smaller than the number of dimensions $d$ ($n<(d+1)$), the covariance matrix will be singular \cite{bull2019outlier}. This can be a serious problem since vibration data are often high dimensional. In damage detection, the curse of dimensionality has motivated the use of ensemble methods to robustly estimate covariance \cite{bull2019outlier,dervilis2015robust}. 

\subsection{Normal-Condition Alignment} 

\begin{figure}[t!]
     \centering
     \includegraphics[width=0.45\textwidth]{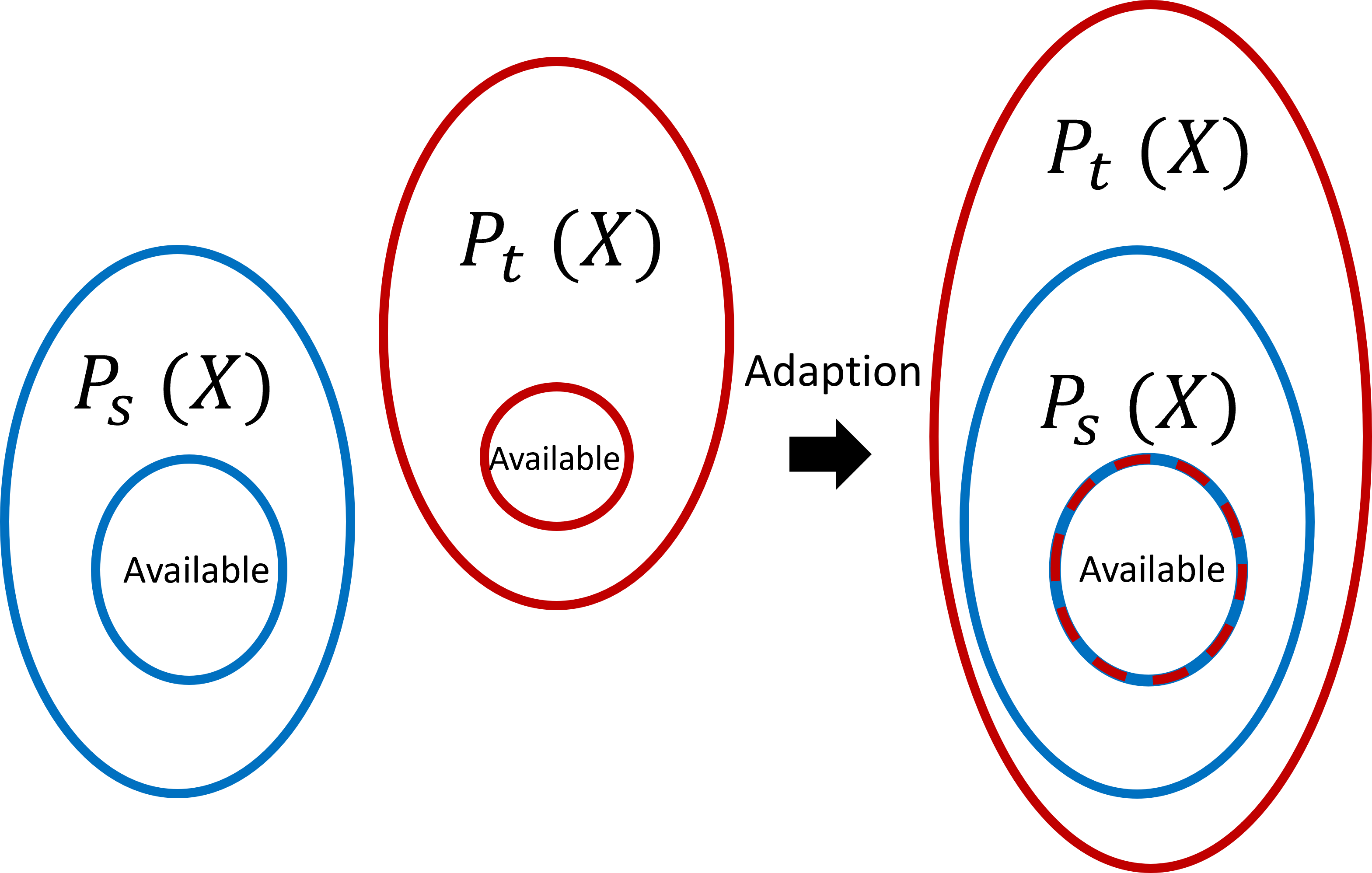}
     \caption{Demonstration of aligning all the available data in the context of partial DA or with class imbalance. The source and target, shown in blue and red respectively, have available data pertaining to a subset of the underlying distribution, shown by the inner circle. After alignment the underlying distributions are not well aligned because the target data is representative of a smaller subset of the underlying distribution.}
     \label{fig: cartoon2}
\end{figure}

SHM datasets are likely to have class imbalance; thus, current SA methods may not be robust to SHM problems. For example, given two bridge datasets it is unlikely that both bridges will have the same quantity of available data from every damage-state and environmental condition. It is also unrealistic to assume the target will contain data from each health-state in the source -- motivating the application of partial DA. As such, their statistics summarise different behaviours and aligning the domains based on these moments will not align the underlying distributions, leading to negative transfer, illustrated in Figure \ref{fig: cartoon2}. Similarly, prominent DA methods typically only minimise the marginal data distributions \cite{Zhuang2021}, which may also result in a subset of classes in the target being aligned to match the distribution of the entire source.  Thus, \emph{normal-condition alignment} (NCA) is proposed, which aims to reduce the risk of aligning data generated via different processes, by utilising the assumption that data gathered at the start of a structures operation were generated by the ``normal-condition" -- a common assumption made for novelty detection \cite{Dervilis2015}.
In NCA, the source domain is first standardised via equation (\ref{eq: source stand}) to centre the data and give features equal treatment. The normal-condition of the target domain is then aligned with that of the source by,
\begin{equation}
    \boldsymbol{z}^{(j)}_t=\left(\frac{\boldsymbol{x}^{(j)}_t-\boldsymbol{\mu}_{t,n}}{\boldsymbol{\sigma}_{t,n}} \right)\boldsymbol{\sigma}_{s,n}+\boldsymbol{\mu}_{s,n}
\end{equation}
\noindent where $\boldsymbol{\mu}_{s,n}$, $\boldsymbol{\mu}_{t,n}$ and $\boldsymbol{\sigma}_{s,n}$, $\boldsymbol{\sigma}_{t,n}$ are the means and standard deviations of the normal-condition data for the source and target respectively. If variation between the datasets is assumed to be limited to scale vector $\mathbf{a}$, and translation vector $\mathbf{b}$, the differences can be expressed by, 
\begin{equation}
X_s=\mathbf{a}X_t + \mathbf{b}
\end{equation}
Given each dataset can be expressed as $X=X_{n} \cup X_{d}$ where $X_{n}$, is normal-condition data  and $X_{d}$ is damage-state data, and the $\mathbf{a}$ and $\mathbf{b}$ are affine transformations, it follows that the transformations for the entire domain can be learnt from the subsets $X_{n}$. It is noted that using only a subset of the data reduces the available data to learn the statistics, but the lower-order statistics should be able to be estimated with a limited sample size.   

The main aim of DA is to adapt the underlying distributions so that a given health-state $c$ from each domain follows the same conditional distribution $P_s(Y=c|X)=P_t(Y=c|X)$. Given that the data distributions for a finite sample will contain biases because of class imbalance and differences in the label spaces, this task is challenging. Previous SA and typical DA approaches would naively attempt to align the marginal distributions, which would not correctly align the underlying distributions, shown in Figure \ref{fig: cartoon2}. Therefore, to align the underlying distributions, a subset of the source label space must be chosen, and the quantity of data in each class should be balanced. Explicitly aligning the marginal distributions of data which are believed to have been generated by the normal-condition is a low-risk way of selecting data corresponding to the same label space. This strategy is in contrast to other approaches, such as automatic weighting or sample selection procedures based on domain discriminators, which may unpredictably select data from different health-states \cite{Li2020}. 

Furthermore, reweighting classes to address class imbalance is challenging as labels are not available in the target. However, if the data are limited to normal-condition data, it can be assumed that variations are related to measurable environmental and operational conditions (EoVs), such as atmospheric temperature, traffic loading, or wind speed etc \cite{cross2013long}. These measurements could allow for further sample selection such that the normal-conditions of each domain contain data from more similar EoVs\footnote{ It should be noted that these measurements will give an indication of what data are affected by certain EoVs, but they will not guarantee the data was generated by a given process.}.

\begin{figure*}[h!]
     \centering
     \begin{subfigure}[b]{0.45\textwidth}
         \centering
         \includegraphics[width=\textwidth]{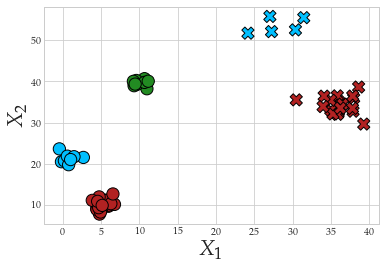}
         \caption{  }
         \label{fig:x1}
     \end{subfigure}
     \hfill
     \begin{subfigure}[b]{0.45\textwidth}
         \centering
         \includegraphics[width=\textwidth]{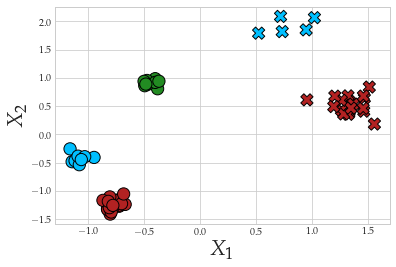} 
         \caption{  }
         \label{fig:x2 }
     \end{subfigure}
      \begin{subfigure}[b]{0.45\textwidth}
         \centering
         \includegraphics[width=\textwidth]{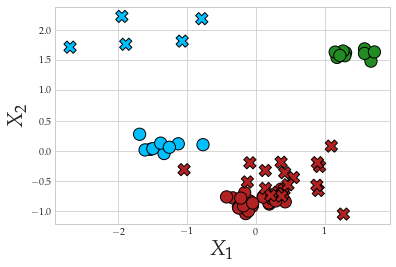}
         \caption{  }
         \label{fig:x3}
     \end{subfigure}
     \hfill
     \begin{subfigure}[b]{0.45\textwidth}
         \centering
         \includegraphics[width=\textwidth]{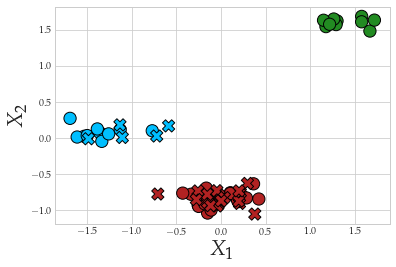} 
         \caption{  }
         \label{fig:x4 }
     \end{subfigure}
     \caption{ Demonstration of aligning a toy example, panel a, presenting a partial DA problem, consisting of three Gaussian clusters in the source, and two in the target. Panel b gives the result of N-standardisation, c A-standardisation and d NCA. The source and target data are represented by (\protect$\medcircle$) and (\protect$\times$) respectively; Classes 0,1 and 2 are depicted in red, blue and green respectively.}
     \label{fig:normnorm}
\end{figure*}

The advantages of aligning the domains based on the normal-condition are demonstrated in Figure \ref{fig:normnorm}. Figure \ref{fig:normnorm}a presents the original toy problem, consisting of a source domain with three Gaussian clusters and a target with two classes. Hence, the problem is a partial DA scenario and there are also differences in the class imbalance between the available classes, with 20 samples in Class 0 (shown in red) for both domains, but with 8 and 4 samples in source and target respectively for Class 1 (shown in blue). Figure \ref{fig:normnorm}b presents N-standardisation, showing that is provides no adaptation.  In Figure  \ref{fig:normnorm}c, A-standardisation has misaligned the standard deviation and mean of the two classes in the target to the three in the source. NCA attempts to address this issue by only considering data from the normal-condition, which in this case aligns the correct classes because the red and blue classes have a similar structure in both domains, shown in Figure  \ref{fig:normnorm}d.

\subsubsection{Normal-Correlation Alignment}
CORAL may also be prone to negative transfer under class imbalance, as accurately estimating the global correlation would be challenging. A modification of CORAL could exploit information in the correlation between the normal data -- Normal-Correlation Alignment (NCORAL). The first step of NCORAL is to apply NCA; correlation alignment is then given by,
\begin{equation}
\underset{A}{min}||\hat{C}_{s,n}-C_{t,n}||_F^2=    \underset{A}{min}||A^TC_{s,n}A-C_{t,n}||_F^2
\end{equation}
where $C_{s,n}$ and $C_{t,n}$ are the correlations of the normal-condition in the source and target respectively. NCORAL extends the assumption that the domains differ by a scale and translation made by NCA, by also considering a rotation.

It is noted that NCORAL learns the correlation from a subset of the entire data, so may have additional issues relating to the curse of dimensionality; this issue may be addressed by ensemble methods for high-dimensional features \cite{Dervilis2015}.

\section{Case Study: Numerical Three-Storey Population}

In SHM, the application of supervised and semi-supervised models is challenging in practical scenarios because label information is often sparse and expensive to obtain \cite{FarrarC.R.CharlesR.2013Shm:}. This section presents a numerical case study, demonstrating the ability of SA to transfer label information to facilitate damage localisation with a limited quantity of data and no labels in the target domain. A number of SA methods are applied -- A-standardisation, CORAL, NCA and NCORAL  -- and these are benchmarked against, N-standardisation (showing the result of applying traditional SHM methods to a population) and a range of DA methods that encompass the general DA approaches used in prior work -- transfer component anaylsis (TCA) \cite{JialinPan2011}, balanced distribution adaptation (BDA) \cite{Wang2017}, the geodesic flow kernel (GFK)\cite{Gong2012} and the domain adversarial neural network (DANN) \cite{Ho2002}.

\subsection{Simulation}


The population consists of two shear structures modelled as $3$DoF lumped--mass models (calculated following the approach in \cite{Gardner2020}). The masses of each DoF were assumed to be a rectangular volume, representing a floor, parameterised by a length $l_m$, width $w_m$, thickness $t_m$, and density $\rho$, with the density sampled from a Gaussian distribution to represent manufacturing variation. The masses were assumed connected by four cantilever beams in parallel, so stiffness is given by $k=4k_b$, where stiffness of each beam was found as the tip stiffness of a cantilever beam, $k_b=\frac{3EI}{l_b^3}$. The elastic modulus $E$ was also drawn from a Gaussian distribution for each sample to introduce variability. Damping $c$ was not derived from a physical model, instead it was, drawn from a Gamma distribution directly.

Damage at a given storey was modelled as an open crack on one of the four beams, located at the midpoint of the beam. It was modelled as a reduction in stiffness as in \cite{Christides1984}; thus, $k=k_d+3k_b$, where $k_d$ is the stiffness of a damaged cantilever beam.

Having obtained the parameters of the model, the damped natural frequencies $\omega_{d}$ were calculated by solving the eigenvalue problem; the first three were used as features, $X \in \mathbb{R}^{n \times 3}$.

Material properties and geometry for each structure are detailed in the Appendix. Data were simulated for the normal-condition and three damage classes, representing damage located at each DoF. For the source structure, 200 samples were collected for each class and labels were assumed known $\{\mathbf{x}_{s,i},y_{s,i}\}_{i=1}^{n_s}$, where $n_s=800$. In the target structure, 100 samples were collected for each class $\{\mathbf{x}_{t,j}\}_{j=1}^{n_t}$, where $n_t=400$. The target labels were assumed to be unknown for all classes apart from the normal-condition. In addition, a separate test target dataset was generated via the same procedure as the training set, i.e. $\{\mathbf{x}_{test,j},\mathbf{y}_{test,j}\}_{j=1}^{n_{test}}$, where $n_{test}=400$.  

In this paper, the case studies focus on vibration-based SHM, but other features could be utilised. It is interesting to note that previous work has applied DA technologies to the similar field of non-destructive testing (NDT) with ultrasound data \cite{gardner2020machine, van2021domain}, and both studies find that normalisation or batch normalisation that adapts the domains has a key role in achieving transfer.

\subsection{Comparison}

To demonstrate that SA can robustly transfer label information between structures, the labels for the three discrete damage locations, corresponding to each DoF, and normal-condition in the source were transferred to the target to facilitate damage localisation for the same health-states, without using target labels. Four SA methods were considered; alignment via A-Standardisation given by equations (\ref{eq: source stand}) and (\ref{eq: target stand}), CORAL, NCA and NCORAL. In addition, to benchmark these methods against traditional SHM (no DA), N-Standardisation (i.e. calculating the statistics from $X=X_s \cup X_t$) was applied. After alignment, a k-nearest neighbours classifier (KNN), with 1 neighbour, was learnt on a source training set and used to classify data in a target test set, although any appropriate classifier can be applied following SA, a KNN is used here, because if the source and target distributions are well aligned, data should be close in Euclidean distance. Furthermore, SA is compared to some prominent DA approaches, using N-standardisation for pre-processing; these are as follows:

\begin{itemize}[leftmargin=1.5em,itemsep=0em]
    \item \emph{Transfer component analysis }(TCA) \cite{JialinPan2011}: learns a feature space that minimises a nonparametric distribution distance metric, the maximum mean discrepancy (MMD) \cite{Gretton2012}, between the marginal distributions $P(X_s)$ and $P(X_t)$, by MMD-regularised kernel principal component analysis (PCA).
    \item \emph{Balanced distribution adaptation} (BDA) \cite{Wang2017}: extends TCA by also attempting to minimise the MMD between the conditional distributions $P(\mathbf{y}_s|X_s)$ and $P(\mathbf{\hat{y}}_t|X_t)$, where $\hat{y}_t$ are label predictions. Since there are no labels in the target, BDA uses pseudo labels from predictions to assign target samples to a given class and finds the distance between the class-conditional distributions $P(X_s|\mathbf{y}_s)$ and $P(X_t|\mathbf{\hat{y}}_t)$.
    \item The \emph{geodesic flow kernel} (GFK) \cite{Gong2012}: uses the ``kernel trick" to find a projection of infinite subspaces on the Grassmann manifold that bridge the source and target PCA subspaces, which are assumed to lie on a common manifold.
    \item The \emph{domain adversarial neural network} (DANN) \cite{Ho2002}: uses a deep feature extractor to find a domain-invariant space. Taking inspiration from the generative adversarial network (GAN) \cite{Goodfellow2014}, a domain discriminator is trained to distinguish between the source and target. Gradient ascent is performed in the feature extractor to align the distributions; a classifier is simultaneously trained to ensure the feature space is discriminative.
\end{itemize}

These methods include the main forms of feature extraction previously used for DA in SHM -- linear subspaces, kernels and DNNs -- as well as methods that minimise a nonparametric statistical distance metric, a domain discrimination loss (adversarial approach) and the distance along the geodesic.

 Hyperparameter selection via cross-validation is challenging  in unsupervised DA, because labels in the target are assumed unavailable. As such, the MMD-based methods (TCA and BDA), utilise an RBF kernel with the length scale chosen as the median of the pairwise distances \cite{Gretton2012}; the dimension of the feature space is reduced by 1 and the Frobenius-norm regularisation parameter was arbitrarily chosen as 0.1; results are largely insensitive to this value \cite{JialinPan2011}. BDA includes a ``balance factor" to control the contribution of the MMD between the marginal and conditional distributions; this was chosen to be 0.5 as suggested in \cite{Wang2017} for the unsupervised setting. The dimension of the GFK subspace must be 1, since there is a requirement that it is less than half the original dimension \cite{Gong2012}. The classifier learnt on the features found by TCA, BDA and the GFK methods was a KNN, with one neighbour. The architecture of the DANN was chosen from a similar case, given in \cite{Poolenorm}. The DANN is sensitive to the random initial weights, so 100 repeats were run and the mean and one standard deviation are given.

Results are given for a test dataset in the target domain, where the evaluation metric used is the macro-F1 score (see \cite{Gardner2020} for more details).

\subsection{Results: Numerical Three--Storey Population}

\begin{figure*}[h!]
\centering
\includegraphics[width=11cm]{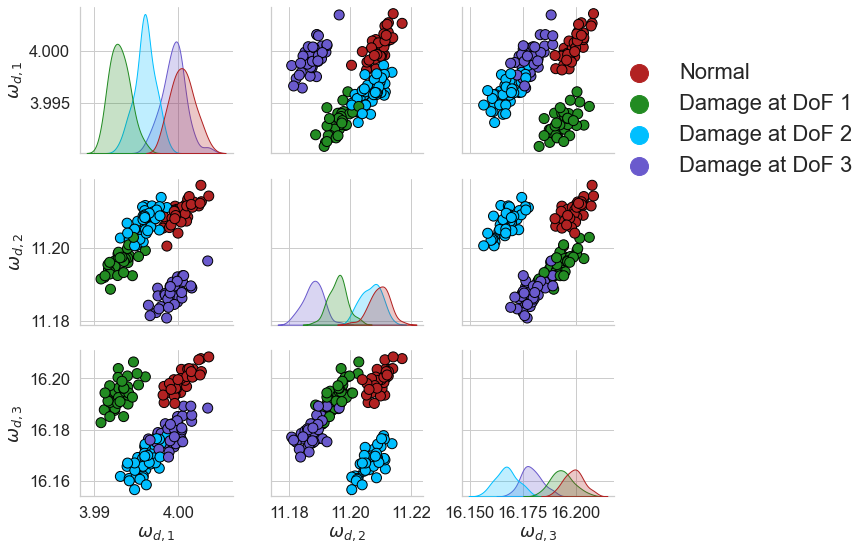}
\caption{Unnormalised damped natural frequencies of the source structure of the numerical  population of three--storey, in Hz. A random subset 20\% the size of the dataset was used for visualisation.}
\label{fig:unnorm source}
\end{figure*}

\begin{figure*}[h!]
\centering
\includegraphics[width=11cm]{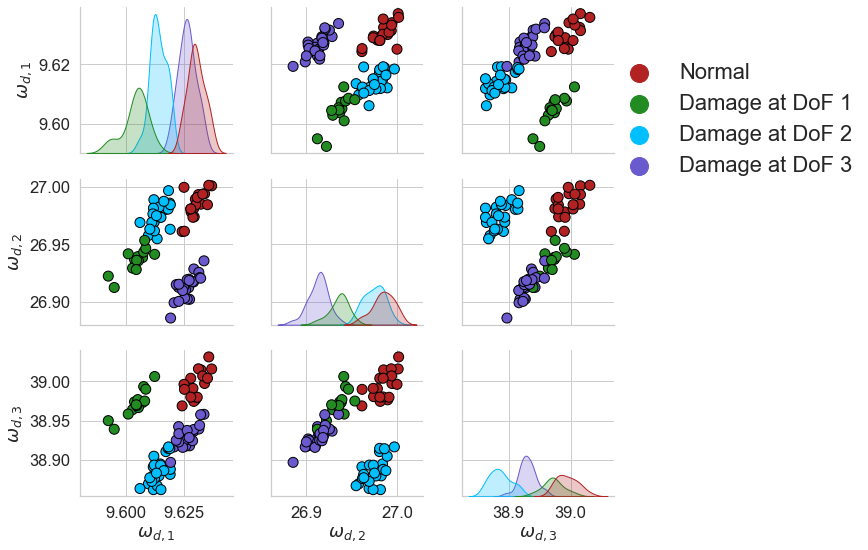}
\caption{Unnormalised damped natural frequencies of the target structure of the numerical  population of three--storey, in Hz. A random subset 20\% the size of the dataset was used for visualisation.}
\label{fig:unnorm target}
\end{figure*}

The unnormalised natural frequencies of the source and target can be found in Figures \ref{fig:unnorm source} and \ref{fig:unnorm target} respectively. The differences between the domains are large, with the absolute values in the target being about a factor of two larger than the source. Estimations of the class data distributions are given by kernel density estimation (KDE) (see \cite{Murphy2014} for more details), shown on the diagonal of each figure.
\begin{figure}[h!] 
     \centering
     \includegraphics[width=0.6\textwidth]{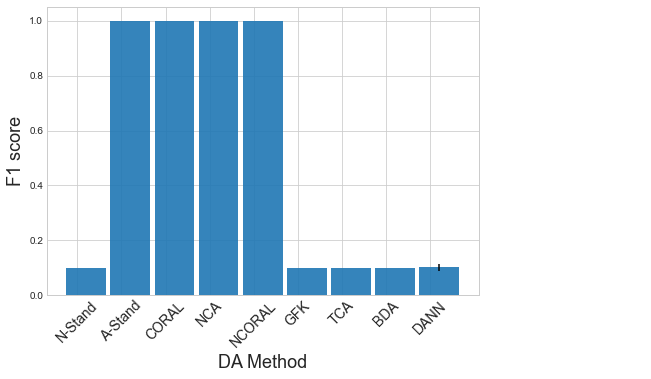}
     \caption{Classification performance of a KNN on the target domain after DA on the numerical three--storey population. The result of the DANN is given as the mean of 100 repeats with one standard deviation shown by a black line.}
     \label{fig:3to3res}
\end{figure}

\begin{figure*}[h!]
\centering
\includegraphics[width=11cm]{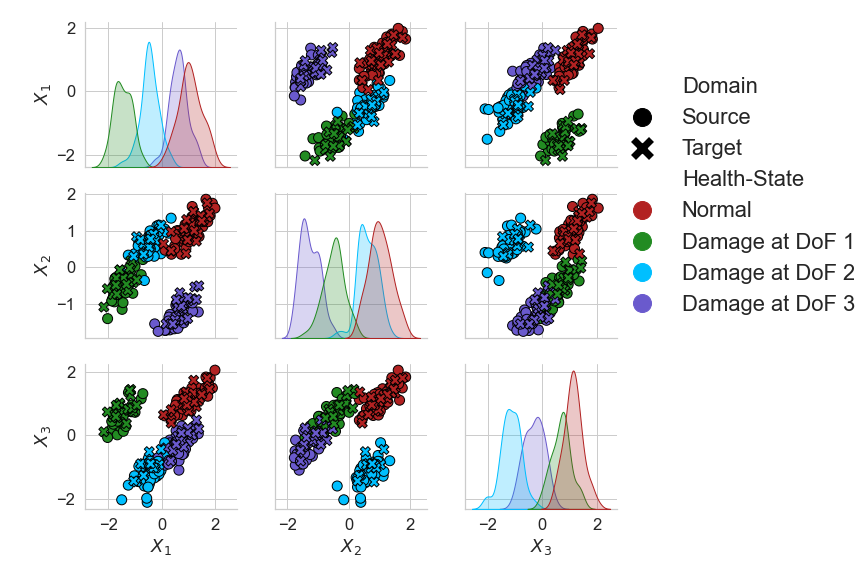}
\caption{ A-standardised features of the numerical three-storey population of structures. The source and target are depicted by ($\medcircle$) and ($\times$) respectively. A random subset 20\% the size of the dataset was used for visualisation.}
\label{fig:3to3feat}
\end{figure*}

Results comparing each DA method can be found in Figure \ref{fig:3to3res}. As expected, the KNN trained on the N-standardised (unadapted) features, which can be considered as naively applying a classifier trained via a traditional SHM approach to another structure, lead to poor generalisation of the source classifier. Following any SA, perfect classification could be achieved. The features given by A-standardisation can be found in  Figure \ref{fig:3to3feat}. It can be seen that even though the two structures have significant differences, the domains have been well aligned, perhaps suggesting differences between similar structures caused by size and material properties may only lead to differences in mean and scale, assuming a linear response. Furthermore, the conventional DA algorithms did not improve classification performance upon N-standardisation. These methods should be able to align both the lower- and higher- order statistics, so this result may suggest that large differences in scale and mean may make learning challenging in conventional DA algorithms. This issue may be related to the quantity of data. It is noted that as the mean and standard deviation are relatively simple to calculate from the data directly, SA could be used to reduce the mean and scale discrepancies before these algorithms are applied, motivating the idea of SA as a pre-processing tool.

\subsection{Results: Partial Domain Adaption with the Numerical Three--Storey Population}

\begin{figure}[h!] 
     \centering
     \includegraphics[width=0.6\textwidth]{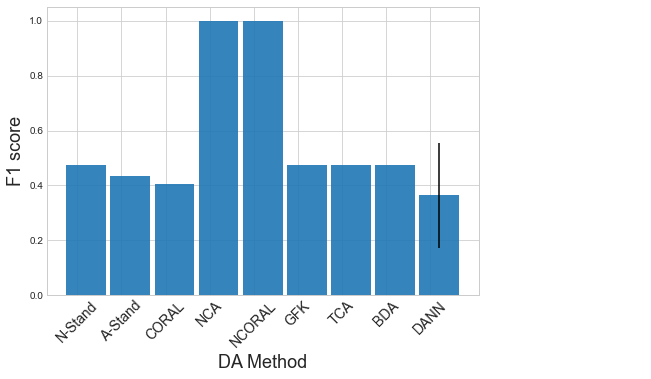}
     \caption{Classification performance of a KNN on the target domain after DA on the numerical three--storey population in a partial DA scenario. The result of the DANN is given as the mean of 100 repeats with one standard deviation shown by a black line.}
     \label{fig:imblanceres}
\end{figure}
\vspace{-0.2cm}

In practical scenarios for PBSHM, it is likely that the amount of data from each health-state and the labels spaces would differ between domains; in such cases, conventional DA is prone to negative transfer \cite{Cao2018a}. Thus, the use of partial DA methods should considered. NCA and NCORAL intrinsically addresses the partial DA problem  by only adapting the normal-condition.
To investigate the robustness of previously applied methods to partial DA and class imbalance, the target domain was downsampled to only include $10$ samples from one damage-state, corresponding to damage on the third storey  -- this is a partial DA problem as the target only has two classes, which are a subset of the four in the source. Therefore, the transfer problem in this case is to transfer the label information from the source structure to the target, given that the target only contains a small quantity of data from one of the three discrete damage locations in the source.

\begin{figure*}[h!]
\centering
\includegraphics[width=11cm]{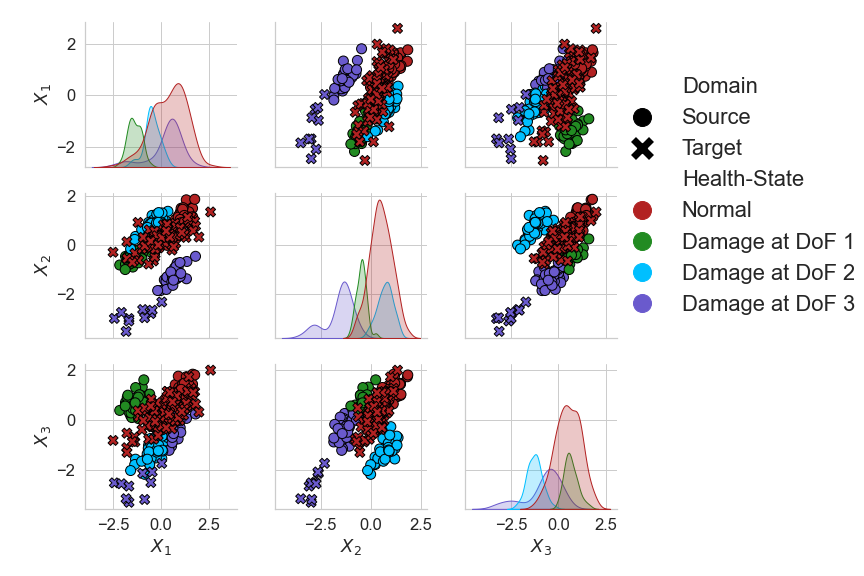}
\caption{ A-standardised features of the numerical three-storey population of structures in a partial DA scenario, with data only from a subset of the source classes in the target. The source and target are depicted by($\medcircle$) and ($\times$) respectively. A random subset 20\% the size of the dataset was used for visualisation.}
\label{fig:3to3Astand}
\end{figure*}

\begin{figure*}[h!]
\centering
\includegraphics[width=11cm]{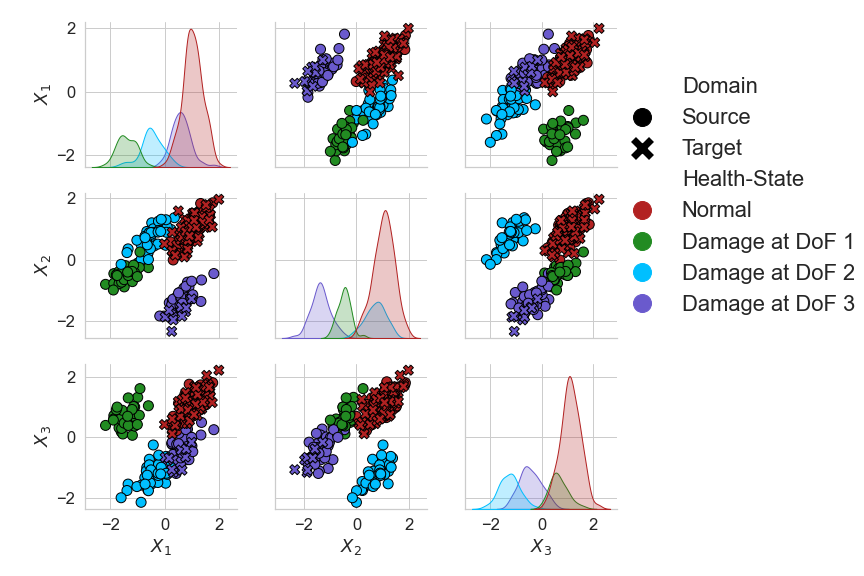}
\caption{ NCA features of the numerical three-storey population of structures in a partial DA scenario, with data only from a subset of the source classes in the target. The source and target are depicted by($\medcircle$) and ($\times$) respectively. A random subset 20\% the size of the dataset was used for visualisation.}
\label{fig:3to3NCA}
\end{figure*}

The results for this case are given in Figure \ref{fig:imblanceres}. As with the previous case, all the conventional DA methods failed to improve generalisation\footnote{The macro-F1 score for N-standardisation is higher than the previous case because there are only two classes in the target domain, so classifying all classes as one class results in a macro-F1 score of 0.5.}. In addition, A-standardisation and CORAL caused negative transfer. Figure \ref{fig:3to3Astand} shows that aligning the global statistics using A-standardisation in this scenario has caused the two available target classes to be spread across the four classes in the source. Figure \ref{fig:3to3NCA} shows that by only aligning the normal-condition statistics, the damage data were also aligned. This result could be achieved without any available damage-state data in the target, allowing damage diagnostics in real time using contextual information from a source domain. 

\section{Case Study: Partial Domain Adaptation with the Z24 and KW51 Bridges}

In a population of real structures, it is more realistic to assume a partial DA setting. This section presents a real population of structure consisting of the Z24 \cite{maeck2003description}, and KW51 bridges \cite{maes2021monitoring}.  This population consists of two partial DA problems that are challenging to solve using previously investigated DA approaches. First, the KW51 bridge was repaired, which changed the response, so to reuse the previous monitoring data DA must be used to realign the response with the pre-repair data. Second, the Z24 bridge dataset contains damage-state data that could be used to further inform a monitoring system if it can be aligned with the KW51. As such, this section demonstrates how NCA and NCORAL can be used to bring these domains into alignment.

\subsection{Datasets}

\begin{figure*}[h!]
\centering
\includegraphics[width=0.8\textwidth]{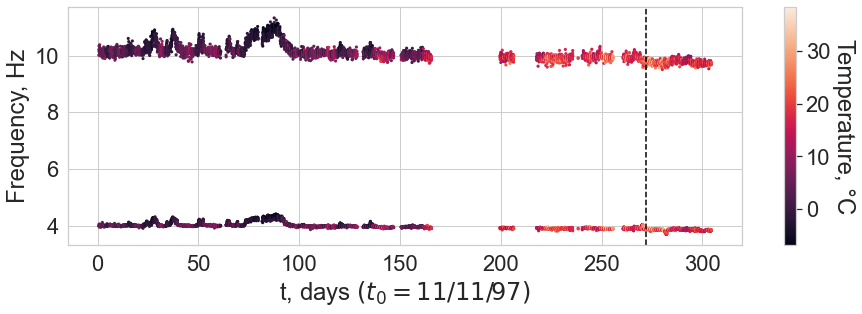}
\caption{The first (bottom) and third (top) natural frequencies of the Z24 bridge dataset. The first instance of damage, commencing on the 10$^{th}$ of August, is indicated by the black line.}
\label{fig:z24 temp}
\end{figure*}

\begin{figure*}[h!]
\centering
\includegraphics[width=0.8\textwidth]{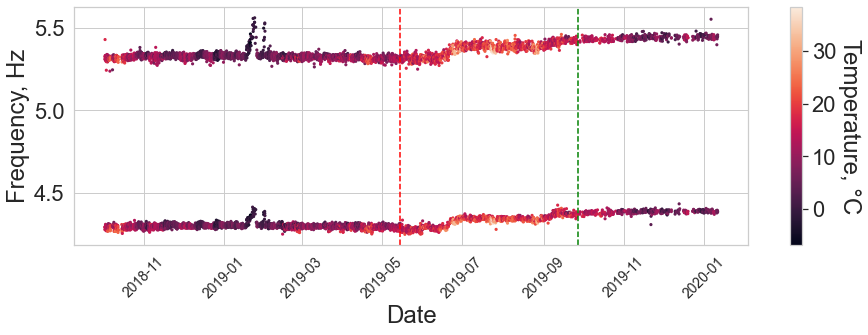}
\caption{The tenth (bottom) and twelfth (top) natural frequencies of the KW51 bridge dataset. The red vertical line indicates the start of the retrofit and the green is the end.}
\label{fig:kw51 temp}
\end{figure*}

The Z24 bridge dataset is well-studied, with data-based approaches being able to identify key events during the monitoring campaign \cite{maeck2001damage,peeters2001one,roeck2003state,teughels2004structural,reynders2010local,Dervilis2015,langone2017automated,bull2019probabilistic,hughes2022risk}. The Z24 bridge was a concrete highway bridge in Bern, Switzerland, which, as part of the SMICES project, was used for an SHM campaign before its demolition in 1998 \cite{maeck2003description}. The first four natural frequencies were found via operational modal analysis (OMA), from the collected acceleration responses. Small-scale damage was introduced by lowering the pier incrementally on the 10$^{th}$ of August 1998, before more severe damage occurred, starting with the failure of the concrete hinge on the 31$^{st}$ of August 1998. For a more in-depth overview of this dataset see \cite{maeck2001damage}.

The KW51 bridge is a steel bowstring railway bridge in Leuven, Belgium. A monitoring campaign  was carried out between 2018 and 2019 during 15 months. The acceleration responses were used to obtain the first 14 natural frequencies via OMA. During the monitoring campaign, each diagonal member was retrofitted with a steel box to strengthen the design of the bridge, with the retrofit beginning on the 15$^{th}$ of May 2019 and completed on the $27^{th}$ of September 2019. Novelty detection of the retrofit has been successfully demonstrated using robust PCA and linear regression trained on the pre-retrofit data \cite{maes2022validation}. For a full description of the dataset, the reader is referred to \cite{maes2021monitoring}. 

Even though the two bridges differ significantly by design, a subset of the natural frequencies can be chosen where there are similarities in the modal response of the bridges. Specifically, the first and third natural frequencies of the Z24 bridge and the tenth and twelfth natural frequencies of the KW51 bridge correspond to vertical bending modes of the deck and have to close the same nodal pattern (for visualisation of the mode shapes see, \cite{maeck2003description} and \cite{maes2021monitoring} for Z24 bridge and KW51 bridge respectively). The corresponding natural frequencies are visualised in Figures \ref{fig:z24 temp} and \ref{fig:kw51 temp}; it can be seen that both bridges experience stiffening effects because of below-freezing conditions and the Z24 bridge dataset contains additional information corresponding to damage.  Therefore, both bridges can be assumed to have two normal-conditions (normal ambient and low temperature). In addition, it is clear that the KW51 is stiffened by repair, shown by the increase the frequencies for similar temperatures, meaning that the pre- and post-repair states correspond to two domains.

Within the two datasets there are three domains. Clearly, the response between the Z24 and KW51 bridges are different and require some form of DA for knowledge transfer. In addition, following repair the response of the KW51 bridge changed. This phenomenon has previously been investigated; it was found for pre-repair data to be used to predict the health-state for the post-repair structure, it must be realigned via DA \cite{Gardner2020b}. Thus, to perform future predictions on the KW51 bridge post-repair using the Z24 and KW51 bridge pre-repair data two partial DA problems must be addressed:
\begin{enumerate}
    \item Align the pre- and post-repair data so that the KW51 bridge data can be treated as one domain. This problem is defined as one of partial DA, because there are data from the ambient and low temperature normal-conditions in the pre-repair state, but only the ambient normal-condition in the post-repair state.
    \item Align the KW51 and Z24 bridge data. This problem is one of partial DA, because there is an additional class in the Z24 bridge dataset relating to damage on the deck.
\end{enumerate}

Using previously investigated DA algorithms, this problem would be challenging to solve. The methods that find a subspace (TCA, BDA, the GFK etc.) will reduce the dimension of the original space to unity when adapting the pre- and post-repair domains, and then they could not be used to adapt the KW51 to the Z24 bridges, as the dimension cannot be further reduced\footnote{Even if there were more features available, the feature reduction performed in step one may make the problem a heterogeneous DA problem, where the feature spaces differ in dimension.  Heterogeneous DA is considered more challenging and requires specialised technologies \cite{Weiss2016}. }. In addition, the DNN-based methods often require labels for a classification task in the source. This classification task maintains the discriminative information in the domains. Although, temperature data are available for each bridge, there are no ground-truth labels that indicate the stiffening effects caused by low temperatures. Therefore,  there are only noisy labels to maintain discriminative information between the ambient and low temperature normal conditions. Even if these labels are used, there are only 110 samples corresponding to low temperatures in the KW51 bridge dataset, so if an unbiased subset of data is chosen there would be 220 samples in the pre-repair source domain, which is likely too small to train a DNN. As such, this section demonstrates that NCA and NCORAL can be used to align the pre- and post- repair states of the KW51 before NCORAL is used to align all the data of the KW51 to the Z24 bridge datasets.

\vspace{-0.1cm}

\begin{figure*}[h!]
     \centering
     \begin{subfigure}[b]{0.45\textwidth}
         \centering
         \includegraphics[width=\textwidth]{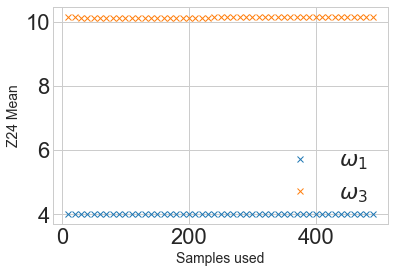}
         \caption{ }
         \label{fig:x5}
     \end{subfigure}
     \hfill
     \begin{subfigure}[b]{0.45\textwidth}
         \centering
         \includegraphics[width=\textwidth]{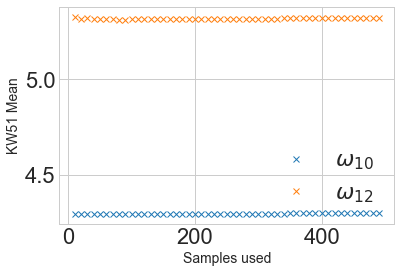} 
         \caption{ }
         \label{fig:x6 }
     \end{subfigure}
      \begin{subfigure}[b]{0.45\textwidth}
         \centering
         \includegraphics[width=\textwidth]{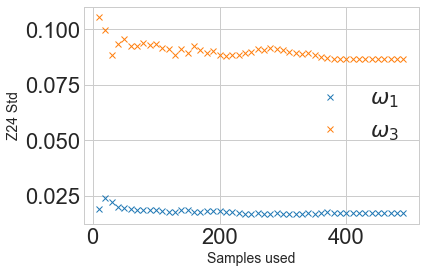}
         \caption{ }
         \label{fig:x7}
     \end{subfigure}
     \hfill
     \begin{subfigure}[b]{0.45\textwidth}
         \centering
         \includegraphics[width=\textwidth]{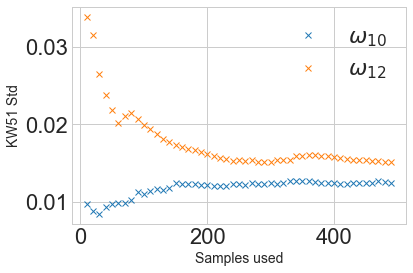}
         \caption{}
         \label{fig:x8 }
     \end{subfigure}
     \caption{Sensitivity analysis for calculating the means and standard deviations for the Z24 and KW51 bridges for a varying sample size. Panels a and b presents the means for the Z24 and KW51 bridges;  panels c and d give the standard deviation. }
     \label{fig:sens}
\end{figure*}

\subsection{Domain Adaptation and Clustering}

\begin{figure*}[h!]
     \centering
     \begin{subfigure}[b]{0.6\textwidth}
         \centering
         \includegraphics[width=\textwidth]{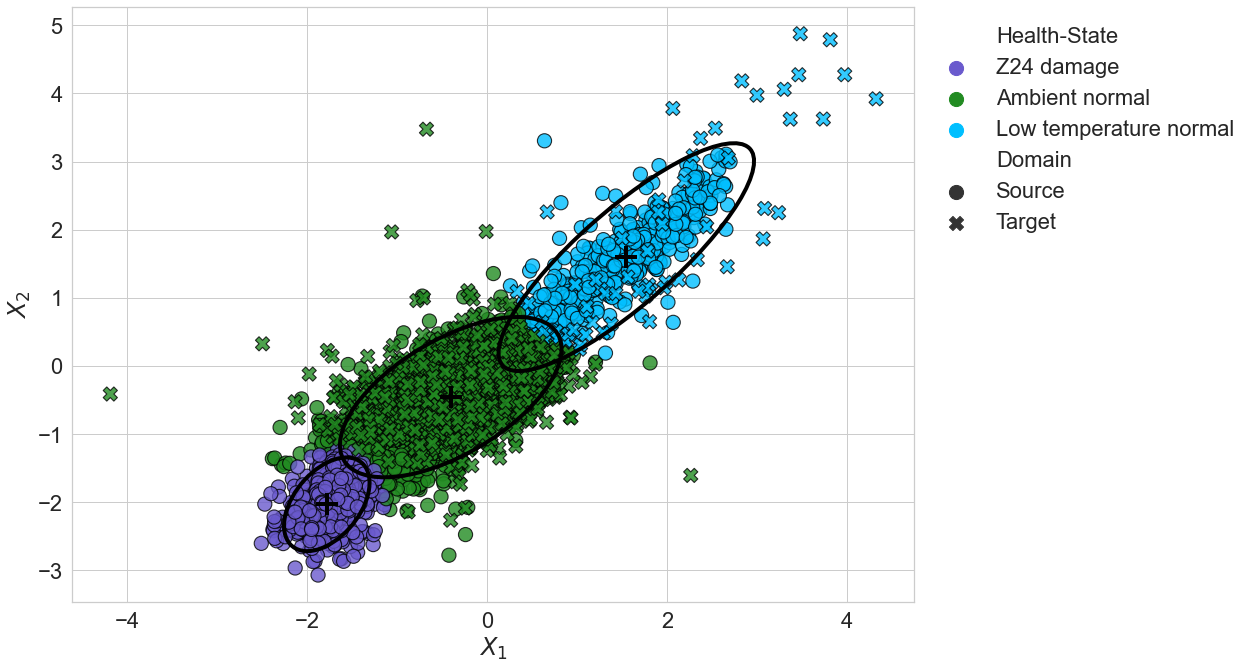}
         \caption{}
         \label{fig:x9}
     \end{subfigure}
     \hfill
     \begin{subfigure}[b]{0.8\textwidth}
         \centering
         \includegraphics[width=\textwidth]{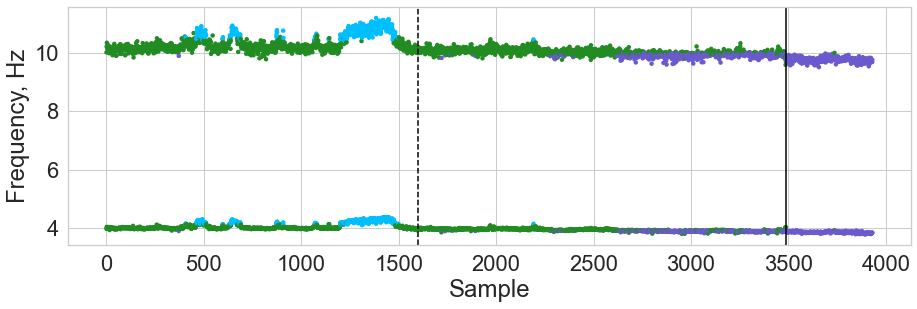}
         \caption{}
         \label{fig:x10}
     \end{subfigure}
     \begin{subfigure}[b]{0.8\textwidth}
         \centering
         \includegraphics[width=\textwidth]{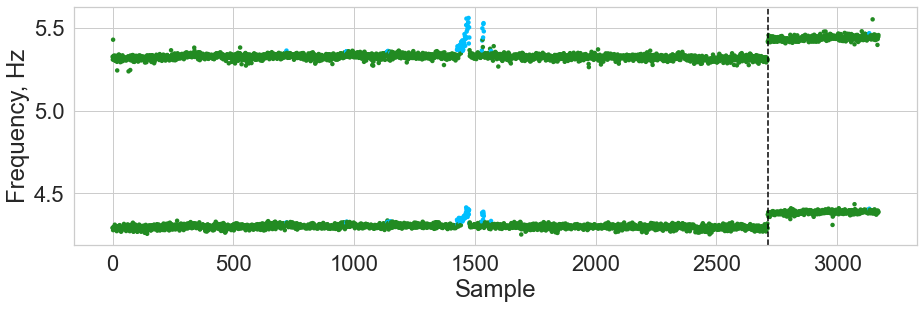}
         \caption{}
         \label{fig:x11}
     \end{subfigure}
     \caption{Unsupervised GMM predictions on the Z24 and KW51 bridge datasets. Panel a gives the comparison of the two features after alignment using NCORAL, showing the four Gaussian components identified ($\mu$ ($+$) and $2\sum$ (----)); the Z24 bridge samples are denoted by ($\medcircle$) and the KW51 by ($\times$). Panels b and c gives the predicted classes on the unadapted Z24 bridge ($\omega_1$ and $\omega_3$) and KW51 bridge ($\omega_{10}$ and $\omega_{12}$) natural frequencies against sample point. }
     \label{fig:gmm}
\end{figure*}

Initially, a sensitivity analysis was conducted to evaluate the quantity of data required in each domain for SA. The mean and standard deviations were calculated for each structure with varying sample size, starting at 10 increasing to 500 samples, with a 10 sample step size. The results of the sensitivity analysis are given in Figure \ref{fig:sens}. It can be seen that the mean can be accurately estimated with very limited data, suggesting transfer could be possible between real structures if the differences are mostly summarised by the mean. As expected, the standard deviations required more data to be accurately estimated, particularly for the KW51. 

The first step was to adapt the pre- and post- repair KW51 data. Thus, NCORAL was applied, learning the statistics using the first 200 data from the pre- and post- repair state, which correspond to similar temperatures. Given the sensitivity analysis (Figure \ref{fig:sens}d), this sample size should correspond to a robust estimation of the statistics whilst only selecting data from a short period after inspection to increase the likelihood that it was generated by the normal-condition.

Once adapted, it was assumed that the pre- and post-repair data could be treated as a single domain. Thus, to align the Z24 and KW51 bridge datasets, NCORAL was applied. As can be seen in Figures \ref{fig:z24 temp} and \ref{fig:kw51 temp}, both bridges experience stiffening effects because of freezing temperatures, so a prior assumption is that, within both datasets, normal-condition data can be split into two classes, pertaining to ambient ($T>0 \degree C$) and low temperatures ($T<0 \degree C$). Labels for these effects are not explicitly known, but temperature data gives an indication as to which data correspond to these effects. As such, data where $T>0 \degree C$ was considered ``ambient normal-condition" and $T < 0 \degree C$ as ``low temperature normal-condition". Since there is significantly more ambient normal-condition data in both domains, the temperature measurements were used for sample selection. For the Z24 bridge, based on the sensitivity analysis (Figure \ref{fig:sens}c), 400 samples were selected from each normal-condition class, with the low temperature normal-condition being randomly selected from the subset where $T<0 \degree C$. In the KW51 dataset, there are only 110 samples corresponding to $T<0 \degree C$; so that the dataset is unbiased 110 samples are used from each normal-condition. Based on the sensitivity analysis (Figure \ref{fig:sens}), this sample size only underestimates the tenth natural frequency by approximately by 4\% and overestimates the twelfth natural frequency by 7\%. The KW51 bridge is adapted by only considering a subset of data from the first 72 days of its monitoring campaign, which can safely to be assumed to be operating in it's normal-condition.

To demonstrate that after alignment information between the bridges can be shared, an unsupervised GMM is learnt on the aligned features. Here an unsupervised model is utilised, since ground-truth labels are unknown, but any appropriate model could be applied after SA. The prior assumption is that there are three classes within datasets, the ambient and low temperature normal-conditions and, the Z24 bridge damage, so a three-component model was implemented. Since, ambient normal data are more abundant, to reflect the prior assumption that normal-condition data is split between ambient and low temperature conditions, the temperature data were used to downsample data corresponding to ambient conditions ($T>0 \degree C$) \footnote{This assumption could also be enforced with Bayesian priors.}.

The aligned features and the predictions made by the GMM can be found in Figure \ref{fig:gmm}. It can be seen in Figures \ref{fig:gmm}b and c, that the ambient normal-conditions of the Z24 bridge, pre- and post-repair KW51 bridge are well aligned, as well as, the low temperature normal-condition of the Z24 bridge and pre-repair KW51 bridge. The feature space has maintained physical interpretability (Figure \ref{fig:gmm}a), an aspect of SA that could be useful for mitigating the risk of negative transfer. The stiffening effect caused by low temperatures engenders
 an increase in each feature (in blue, where $T<0\degree C$). In addition, the stiffening reduction caused by damage in the Z24 bridge can be seen by a reduction in each feature (in purple). 

One of the main advantages of aligning the population of bridges is that damage-state data from a source could be used to further inform a damage detector for the target. In Figure \ref{fig:gmm}, it can be seen that the normal-condition data for the KW51 lies on the boundary with damage in the Z24 bridge dataset, but no data are misclassified as showing damage\footnote{Note, the Z24 bridge damage labels were not used to learn this GMM, as the main aim is to show the domains are well aligned by SA. A supervised or semi-supervised model could be used to better define this boundary.}. This result motivates the idea of using damage to inform a novelty detector to reduce the number of false positives that occur when the threshold is defined based on the normal-condition of the target alone. Unfortunately, there is no damage in the KW51 to confirm this finding.

This case study also illustrates a trade-off between selecting features that are transferable and damage sensitive, shown by Figure \ref{fig:gmm} b where some normal-condition data of the Z24 bridge dataset are clustered with damage. It can be seen in Figure \ref{fig:gmm} a, that this is largely because damage is masked by ambient Z24 bridge data. In previous studies however, is has been demonstrated that using all four available frequencies allows for damage to be discriminated \cite{Dervilis2015}, but the most damage sensitive natural frequency in the Z24 bridge dataset (the second natural frequency) was not used since the mode shapes indicate there is low physical similarity with any of the modes in the KW51 bridge. This trade-off may be influenced by the similarity of the population. In this population, a concrete box-girder highway bridge is used to transfer information to a steel bow-string railway bridge, with the two structures having differences in material properties, geometries and connectivity. Thus, since the aim of DA here is to find a feature space where future health-state data could be shared, the features should share physical similarity, such that it is believed that the same physical phenomena in each structure would correspond to similar parts of the feature space. This objective justifies the approach of selecting only frequencies with strong modal correspondence, but the dissimilarity between the bridges means that only  two frequencies were deemed to be sufficiently similar. In a population where structures are more similar it is reasonable to expect that a larger proportion of modes would be similar, reducing the severity of this trade-off; for example, two nominally identical wind turbines would be expected to have a larger proportion of similar modes.

Despite the masking effects, the aim of this case was to demonstrate that NCA and NCORAL can align the feature spaces of two structures in a partial DA scenario, as well as address changes to the structural response caused by repair by only using limited normal-condition data gathered in a short period after inspection. This objective has been achieved using a small quantity of inexpensive data, i.e. normal-condition response and temperature data. The trade-off between transferable and damage sensitive features is a topic for further research. 

\section{Case study: Statistic Alignment as Pre-processing}




The previous case studies have demonstrated that exclusively aligning the lower-order statistics via SA can facilitate knowledge transfer. In fact, if the underlying data distributions are similar enough, SA can completely align the domains, as in the three-storey case study. 

It is noted in contexts where the datasets are similar and SA is sufficient for adaptation, applying further adaptation may add an additional risk of negative transfer and lose the physical interpretability of the feature space. The question of whether similarity can be measured before transfer has been considered, with \cite{Gosliga2021} and \cite{Gardner2020} proposing techniques for measuring physical and data similarity between members of a population.

In scenarios where members of the population are less similar, there may be disparity between both the lower- and the higher-order statistics. Thus, it is proposed that SA should be used as pre-processing, aligning the lower-order statistics, such that the nonlinear mapping found by further DA is simplified.

Compared to SA, specifically NCA and NCORAL, prominent DA algorithms may have a number of additional requirements. Data available in each domain should be representative of their respective underlying distributions and abundant enough for nonparametric density estimation. Furthermore, prominent DA methods are prone to negative transfer in partial DA \cite{Cao2018a} and if the target only represents of small subset of the source classes, the risk of negative transfer will be higher \cite{shui2020beyond}. Therefore, further DA should be used when there is reason to assume that there are significant differences in the higher-order statistics and also, when  available information is sufficient in each domain to mitigate the associated risk of negative transfer.


To demonstrate SA as a pre-processing method, another numerical case study is presented. The simulation procedure follows the previous numerical case study, with this case forming a heterogeneous population consisting of 3DoF source and 7DoF target structure, giving a more complex transfer problem (the material properties can be found in the Appendix). Damage was simulated for the first and third storeys in each structure. This case does not investigate partial DA, as prominent DA methods are prone to negative transfer in this scenario so identification of robust partial DA methods to use in conjunction with SA is left for future research. For the source, 400 samples were simulated for the normal-condition and 150 samples for each of the damage-states, $n_s=700$. In the target, 200 samples were simulated for the normal-condition and 75 samples for each of the damage-states, $n_t=350$. Class imbalance between the normal-condition and damage-states is introduced in this way to emulate practical scenarios, where normal-condition data will be more abundant. The transfer learning problem is, therefore, to transfer the label information from the normal-condition, first damage location- and third damage location- in the source structure to the same location in the target structure.

N-standardisation, A-standardisation, CORAL and NCA were each used as a normalisation procedure. Following this step, each of the previously used DA methods were applied, including a KNN on the N-standardised space, with one neighbour, to provide a benchmark for performance gains given by further DA. 


\subsection{Results: Numerical Three-- to Seven-- Storey Population}
\begin{figure*}[h!]
\centering
\includegraphics[width=11cm]{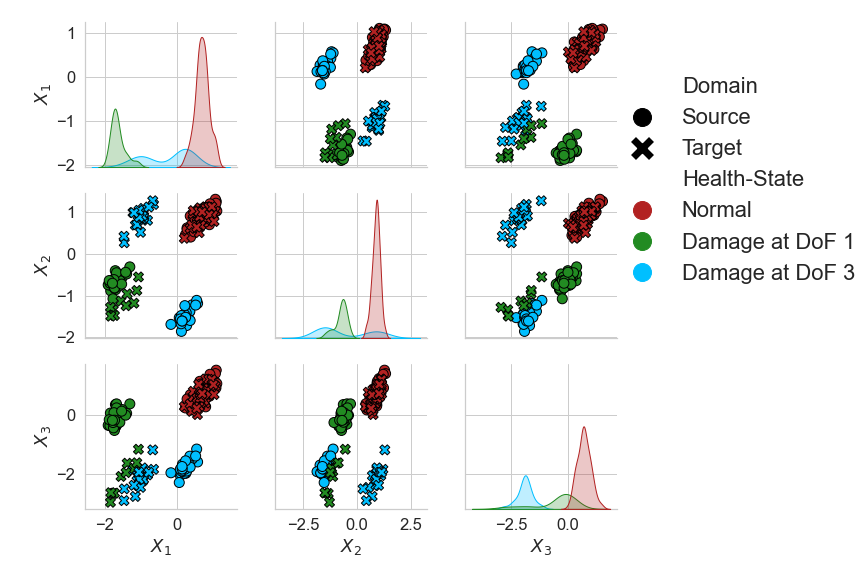}
\caption{ NCA features of the numerical three- to seven-storey population of structures. The source and target are depicted by ($\medcircle$) and ($\times$) respectively. A random subset 20\% the size of the datasets is plotted for visualisation}
\label{fig:NCAnotokay}
\end{figure*}

\begin{figure}[h!] 
     \centering
     \includegraphics[width=0.5\textwidth]{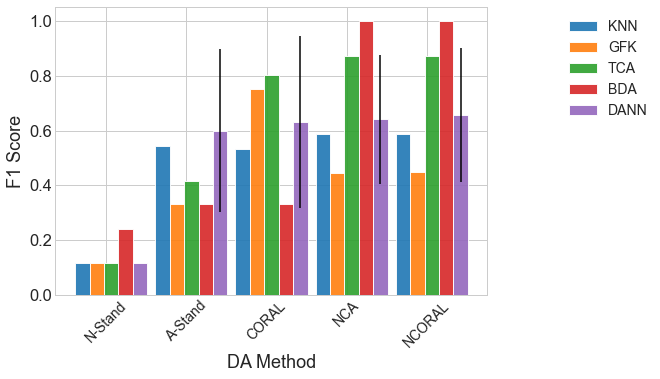}
     \caption{Classification performance of a KNN on the target domain after various SA methods, then DA on the numerical three-- to seven--storey population. The result of the DANN is given as the mean of 100 repeats with one standard deviation shown by a black line.}
     \label{fig:NCAisnorm}
\end{figure}
The NCA features are visualised in Figure \ref{fig:NCAnotokay}. Significant differences in the domains can be seen, including the order of Classes 1 and 2 being flipped between domains in the second mode. Classification results can be found in Figure \ref{fig:NCAisnorm}. Without first performing statistic alignment, none of the nonlinear DA methods achieved adequate adaptation for knowledge transfer. Each of the SA methods alone improved upon the unaligned KNN. Following NCA and NCORAL, excluding the GFK, all DA methods were able to improve upon SA alone. In this case, NCORAL gives the same result as NCA, since correlation of the normal-conditions is the same in both domains. The GFK did result in negative transfer after NCA and CORAL, but this method aligns PCA subspaces with dimension $k$, with the condition it must be under half the dimension of the original space $d$, i.e. $k<d/2$. Thus, in this case $k=1$, which may have not been sufficient to encode enough discriminative information. A-standardisation and CORAL did perform as well as NCA or NCORAL with any method, with the exception of the GFK after CORAL, suggesting that class imbalance effected their performance. These results suggest that even though without SA, the nonlinear DA methods fail to achieve knowledge transfer, they can still provide additional performance gains if SA is applied first, highlighting that it is crucial to consider SA for transfer in SHM.

\section{Conclusions}
\label{sec:conclusion}

PBSHM aims to facilitate more in-depth health-state diagnosis in SHM by sharing information across a population of structures. When utilising information across different structures, whether that be a homogeneous or heterogeneous population, there will be differences between the training and testing distributions. Transfer learning, in the form of domain adaptation, offers a solution to this issue by finding a mapping that aligns the feature spaces. However, previously-implemented DA methods have relied on methods that are data intensive, susceptible to negative transfer under class imbalance and in partial DA, and map data into a latent space that may make it challenging to interpret physical phenomena. SA is proposed as a solution to these challenges, with two methods proposed to address issues regarding class imbalance and partial DA. It is also proposed that SA can simplify the problem for nonlinear methods when used as a pre-processing method. 

Three case studies were presented to evaluate the applicability of SA. The first, a heterogeneous numerical population, found that the differences between linear structural responses caused by material properties and the size of the structure could be removed by aligning the mean and standard deviations. Furthermore, the the main approaches to DA that have been previously studied failed to adapt the domains. The dataset was then downsampled to investigate class imbalance, in a partial DA scenario, suggesting that previous SA approaches may cause negative transfer under class imbalance, whereas NCA and NCORAL are more robust. 

The second case study presented a real heterogeneous population of two bridges -- the Z24 and KW51 bridges. NCORAL was applied, solving two partial DA problems relating to the pre- and post-repair states in the KW51 bridge dataset, as well as adapting the KW51 and Z24 bridge datasets. Adaptation was achieved by using limited data directly after inspection in the KW51 bridge, which was shown to be adequate to identify behaviour of interest using an unsupervised GMM. In addition, the idea of sample selection using measurable EoVs to reduce class imbalance was demonstrated by selecting data  corresponding to above and below freezing temperatures. Intuitively, this population may be considered as a highly-challenging transfer scenario, as it consists of a concrete highway bridge and a steel railway bridge. Thus, the fact that NCORAL could align the domains with such a small dataset demonstrates the potential of aligning normal-condition data. A potential application of this alignment would be to define a more informed threshold for damage detection, using the damage in the Z24 bridge dataset. Such a threshold would result in no false positive in the KW51 bridge dataset despite KW51 bridge normal-condition data being extremely close to boundary for the identified damage cluster in the feature space after alignment.

In dissimilar populations only considering the lower-order statistics may limit the potential performance gains that a nonlinear mapping could provide, so a final case study discusses using SA as a pre-processing method to simplify the DA problem. Even though the nonlinear DA methods utilised should have the ability to align the lower- and higher-order statistics, without first using SA, all of these methods failed to transfer knowledge. This result suggests that SA is a potentially crucial pre-processing method for DA in SHM. NCA and NCORAL were shown to be particularly robust to class imbalance. 

Future work is required for the reliable application of the technologies discussed in this paper. One limitation of the DA identified here, is the trade-off between discriminative and transferable features, motivating the investigation of the required level of similarity between features for robust transfer. In addition, this paper suggests that for similar features, SA can achieve sufficient adaptation to provide more accurate; in-depth diagnostics and in situations where further DA is beneficial, SA is a crucial pre-processing step. Since there is a risk of negative transfer associated with applying DA \cite{Zhang2020}, DA should not be naively applied, so future work is required to identify when it is beneficial and how to reduce the risk of negative transfer in an unsupervised manner. A large part of this work will be concerned with identifying a suitable unsupervised metric to measure similarity between features; for example, physical similarity may be used \cite{Gosliga2021}.  Also, methods used in conjunction with SA for aligning higher-order statistics should also be robust to partial DA problems so the development of partial DA algorithms that do not require large datasets will also be considered. Finally, partial DA has been demonstrated as a useful tool to alleviate the strict assumption that the label spaces are identical, as made by conventional DA. Nevertheless, in some scenarios, the target domain will also contain data pertaining to a health-state that is not in the source, motivating the study of methods that further weaken this assumption; for example, open-DA \cite{Zhuang2021}.

\section*{Acknowledgements}

The authors would like to acknowledge the support of the UK Engineering and Physical Sciences Research Council via grants EP/R006768/1, EP/R003645/1 and EP/R004900/1. Lawrence Bull was supported by Wave 1 of the UKRI Strategic Priorities Fund under the EPSRC Grant EP/W006022/1, particularly the \textit{Ecosystems of Digital Twins} theme within that grant and The Alan Turing Institute.
The authors would also like to thank Dr. Kristof Maes, Dr. Edwin Reynders
and Prof. Geert Lombaert for providing access to the KW51 dataset.

\bibliographystyle{IEEEtran}
\bibliography{ref}
\appendix

\vspace{200cm}
\clearpage
\begin{appendices}
\section{Material Properties}

\begin{table}[h!]
\centering
\begin{tabular}{|l|l|l|l|} 
\cline{2-4}
\multicolumn{1}{l|}{}               & Unit     & Source                     & Target                      \\ 
\hline
Beam geometry~$\{l_b,w_b,t_b\}$     & $mm$     & $\{300,40,8\}$             & $\{160,25,6\}$              \\ 
\hline
Mass geometry~$\{l_m,w_m,t_m\}$     & $mm$     & $\{400,400,40\}$           & $\{300,250,25\}$            \\ 
\hline
Crack geometry~$\{l_{cr},l_{loc}\}$ & $mm$     & $\{20.0,150\}$             & $\{12.5,80\}$                \\ 
\hline
Elastic modulus $E$                 & $GPa$    & $\mathcal{N}(210,1\times 10^{-9})$ & $\mathcal{N}(71,1\times 10^{-10})$  \\ 
\hline
Density~$\rho$                      & $kg/m^3$ & $\mathcal{N}(7800,50)$             & $\mathcal{N}(2700,10)$             \\ 
\hline
Damping coefficient~$c$             & $Ns/m$   & $\mathcal{G}(8,0.8)$       & $\mathcal{G}(50,0.8)$       \\
\hline
\end{tabular}
\caption{ \centering Properties for the numerical 3-storey case study.}
\end{table}

\begin{table}[h!]
\centering

\begin{tabular}{|l|l|l|l|} 
\cline{2-4}
\multicolumn{1}{l|}{}               & Unit     & Source                     & Target                      \\ 
\hline
Beam geometry~$\{l_b,w_b,t_b\}$     & $mm$     & $\{300,40,8\}$             & $\{300,40,8\}$              \\ 
\hline
Mass geometry~$\{l_m,w_m,t_m\}$     & $mm$     & $\{400,400,40\}$           & $\{400,400,40\}$            \\ 
\hline
Crack geometry~$\{l_{cr},l_{loc}\}$ & $mm$     & $\{20.0,150\}$             & $\{20.0,150\}$                \\ 
\hline
Elastic modulus $E$                 & $GPa$    & $\mathcal{N}(210,1\times 10^{-9})$ & $\mathcal{N}(210,1\times 10^{-9})$  \\ 
\hline
Density~$\rho$                      & $kg/m^3$ & $\mathcal{N}(7800,50)$             & $\mathcal{N}(7800,50)$             \\ 
\hline
Damping coefficient~$c$             & $Ns/m$   & $\mathcal{G}(8,0.8)$       & $\mathcal{G}(8,0.8)$       \\
\hline
Storeys                             &          & $3$                                 & $7$                                 \\
\hline
\end{tabular}
\caption{\centering Properties for the three- to seven-storey numerical case study}
\end{table}
\end{appendices}
\pagenumbering{roman}
\end{document}